%% file: main.tex
\documentclass[10pt,twocolumn,letterpaper]{article}

\usepackage{cvpr}              
\usepackage{multirow}
\usepackage{graphicx}
\usepackage{tabularx}
\usepackage{tablefootnote}
\usepackage{threeparttable}
\usepackage{diagbox}
\usepackage{parskip}

\definecolor{cvprblue}{rgb}{0.21,0.49,0.74}
\usepackage[pagebackref,breaklinks,colorlinks,allcolors=cvprblue]{hyperref}

\def\paperID{*****} 
\def\confName{CVPR}
\def\confYear{2025}

\title{Inclusion 2024 $\cdot$ Global Multimedia Deepfake Detection Challenge: Towards Multi-dimensional Face Forgery Detection}

\author{
Yi Zhang\\
\and
Weize Gao\\
\and
Changtao Miao\\
\and
Man Luo\\
\and
Jianshu Li \thanks{Corresponding Author: jianshu.l@antgroup.com}\\
\and
Wenzhong Deng\\
\and
Zhe Li\\
\and
Bingyu Hu\\
\and
Weibin Yao\\
\and
Yunfeng Diao\\
\and
Wenbo Zhou\\
\and
Tao Gong\\
\and
Qi Chu\\
}

\begin{document}
\maketitle
\input{sec/0_abstract}    
\input{sec/1_intro}
\input{sec/2_formatting}
\input{sec/3_finalcopy}
\input{sec/4_solution}

\input{sec/5_discussions}
\input{sec/6_conclusion}
{
    \small
    \bibliographystyle{ieeenat_fullname}
    \bibliography{main}
}


\end{document}

%% file: sec/0_abstract.tex
\begin{abstract}
In this paper, we present the Global Multimedia Deepfake Detection held concurrently with the Inclusion 2024. Our Multimedia Deepfake Detection aims to detect automatic image and audio-video manipulations including but not limited to editing, synthesis, generation, Photoshop,~\emph{etc}. Our challenge has attracted 1500 teams from all over the world, with about 5000 valid result submission counts. We invite the top 20 teams to present their solutions to the challenge, from which the top 3 teams are awarded 
prizes in the grand finale. In this paper, we present the solutions from the top 3 teams of the two tracks, to boost the research work in the field of image and audio-video forgery detection. The methodologies developed through the challenge will contribute to the development of next-generation deepfake detection systems and we encourage participants to open source their methods \footnotemark[1].
\end{abstract}
\footnotetext[1]{https://github.com/inclusionConf/DeepFakeDefenders/}

%% file: sec/1_intro.tex
\section{Introduction}
\label{sec:intro}

With the explosive advancement of AIGC deep synthesis technologies, capabilities in facial deepfake generation have significantly improved, allowing malicious actors to create highly realistic fake faces. In recent years, the misuse of facial deepfake technology has garnered substantial public concern. In real-world digital identity verification scenarios, criminal organizations have employed such technology to compromise facial recognition systems. If your face is substituted in a facial recognition video, that counterfeit clip could potentially be utilized by cybercriminals to exploit your digital accounts. Therefore, enhancing the security of biometric identification necessitates the urgent development of effective facial deepfake detection techniques.

\begin{table*}[ht]
    \centering
    \scalebox{0.8}{
    \begin{tabular}{c|c|c}
    \toprule
         Dataset & Manipulated Modality & The Number of Generation Methods\\
         \midrule
         FaceForensics++ \cite{rossler2019faceforensics++} & Video & 4\\
         \midrule
         Celeb-DF \cite{li2020celebdf} & Video & 1-2 \\
         \midrule
         DiFF \cite{cheng2024diff} & Image & 13\\
         \midrule
         LAV-DF \cite{cai2023lavdf} & Audio-Video & 1\\
         \midrule
         AV-Deepfake1M \cite{cai2023avdf1m} & Audio-Video & 1\\
         \midrule
         \multirow{2}*{\textbf{MultiFF (ours)}}&Image& 81\\
         ~&Audio-Video&100+ \footnotemark[2]\\
         \bottomrule
    \end{tabular}}
    \caption{The comparison with other Deepfakes Datasets.}
    \label{tab:0}
\end{table*}

The effectiveness of deepfake detection methods \cite{miao2021learning,miao2021towards,miao2022hierarchical,miao2023f,zhuang2022towards,zhuang2022uia,tan2022transformer} is highly dependent on the datasets. We investigated deepfake datasets commonly used in academia and industry and recorded the types of forgery methods they used in Table \ref{tab:0}. Although the past few years have seen an increase in publicly available datasets focused on image and audio-visual content manipulations, most of these datasets contain a single or a few generative methods. However, in the realm of deepfake detection, a significant challenge for detectors is to generalize effectively to unseen deepfake sources in real-world scenarios. A conspicuous gap remains in the lack of source-invariant representation exploited from the generator pipeline for forgery image or audio-video detection. This deficiency leads to failures in detecting unknown forgery domains.

\begin{figure*}[ht]
\centering
\includegraphics[scale=0.1]{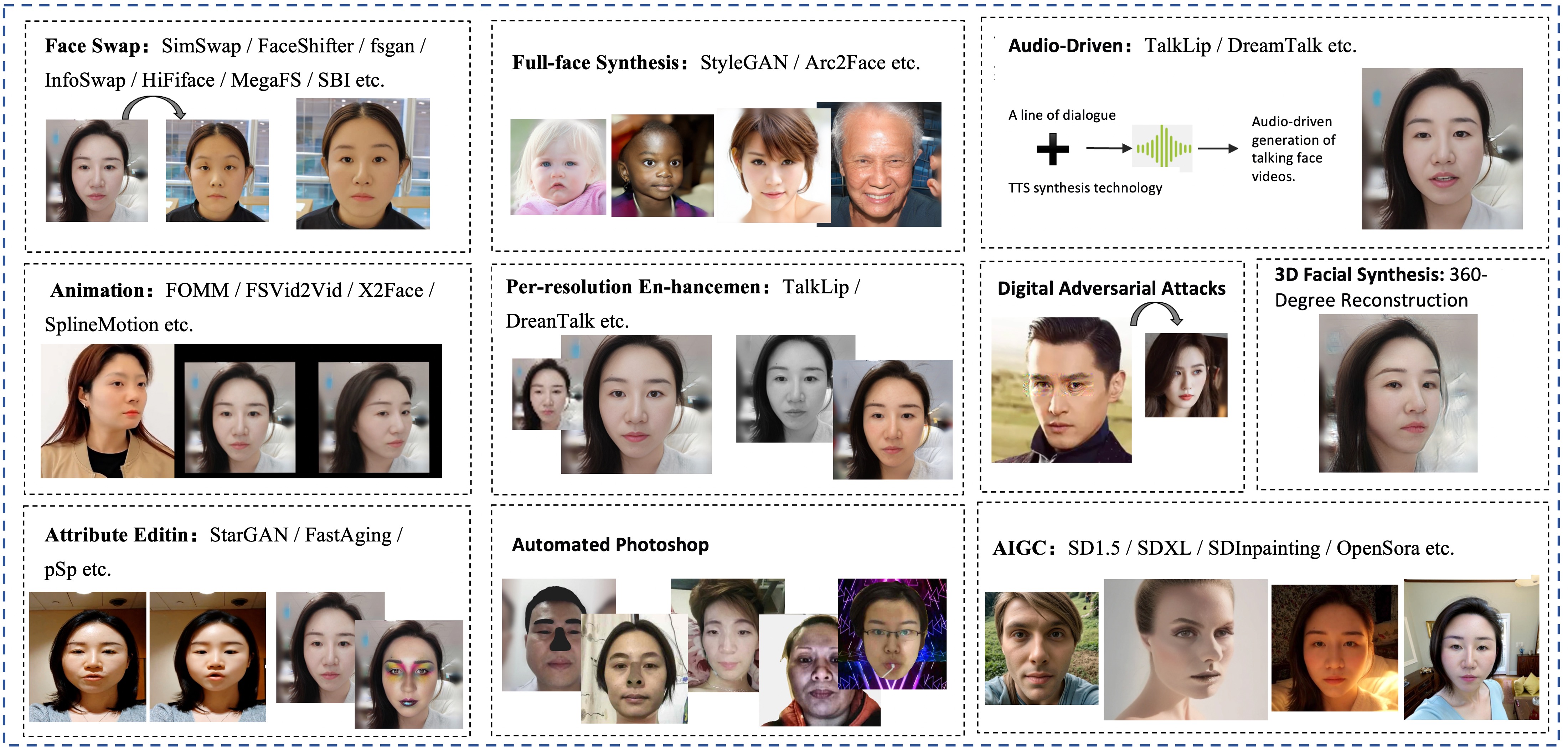}
\caption{The construction of our MultiFF dataset.}
\label{fig:1_1}       
\end{figure*}

To overcome the gap, the Multi-dimensional Facial Forgery (MultiFF) dataset was introduced, providing a large-scale benchmark of images and audio-videos for the task of deepfake detection, which specifically includes two subsets: Multi-dimensional Facial Forgery Image dataset (MultiFFI) and Multi-dimensional Facial Forgery audio and Video dataset (MultiFFV). The images in the MultiFFI dataset are generated by more than 80 atomic generation algorithms. The total generation methods in MultiFFV are more than 100. Based on this dataset, the Global Multimedia Deepfake Detection challenge focuses on the binary classification of deepfake content. The challenge is planned to contribute to improving current detection methods and aims to run as an ongoing benchmarking for the next several years, continually introducing new challenges of deepfake technology to keep pace with its rapid evolution.

The rest of the paper is organized as follows. We will first demonstrate the setup of our challenge, and then present the details of the solutions from the top 3 teams. After that, we will discuss the results from the teams and conclude the challenge.

\footnotetext[2]{There are 40 and 7 generation methods for video and audio modalities respectively, therefore the total number of modalities combination generation methods exceeds 100.}

%% file: sec/2_formatting.tex
\section{Datasets}
\label{sec:datasets}

In our challenge, we released a new diversified fake digital face dataset named MultiFF, which specifically includes two subsets: MultiFFI and MultiFFV, which will be used for the image deepfake detection task in Track 1 and the audio-video deepfake detection task in Track 2 respectively. The MultiFFI \footnotemark[3] dataset contains over 900,000 images which are generated by more than 80 atomic generation algorithms. It is sourced from four diverse facial datasets (CelebA, RFW, CASIA$\_$Webface, and some open online faces) and includes techniques such as face swapping (SimSwap \cite{chen2020simswap}, FaceShifter \cite{li2019faceshifter}, FSGAN \cite{nirkin2019fsgan}, InfoSwap \cite{gao2021information}, etc.), animation (FOMM \cite{siarohin2019first}, ArticulatedAnimation \cite{siarohin2021motion}, etc.), attribute editing (DualStyleGAN \cite{yang2022pastiche}, GPEN$\_$Colorization \cite{yang2021gan}, etc.), full-face synthesis (StyleGAN2 \cite{karras2020analyzing}, StyleGAN3 \cite{karras2021alias}, etc.), super-resolution enhancement (FaceSR, CodeFormer \cite{zhou2022towards}, etc.), and AIGC (SD1.5, SDXL$\_$Inpaiting, etc.), among others. Additionally, it encompasses a variety of facial attack materials, including diverse skin tones and ethnicities, different angles and poses, occlusions (such as glasses, masks, hats, and bangs), a rich array of indoor and outdoor scenes, as well as variations in age and lighting conditions. The total volume of the MultiFFV dataset exceeds 600,000, with facial video sources including VoxCeleb \cite{nagrani2017voxceleb}, CelebV-HQ \cite{zhu2022celebv}, and VFHQ \cite{xie2022vfhq}. The real human audio sources comprise VCTK \cite{liu2019cross}, TalkingHead, and LJSpeech \cite{xu2020lrspeech}. The MultiFFI and MultiFFV in our challenge include over 150 types of image and audio-video forgeries so that the participants have ample design space to model forgery types. The image and audio-video numbers of MultiFFI and MultiFFV in the proposed MultiFF dataset are shown in Table \ref{tab:1} in detail.
\footnotetext[3]{The publicly available portion of the MFFI dataset has been released at https://github.com/inclusionConf/MFFI}

\begin{figure}[ht]
\centering
\includegraphics[scale=0.45]{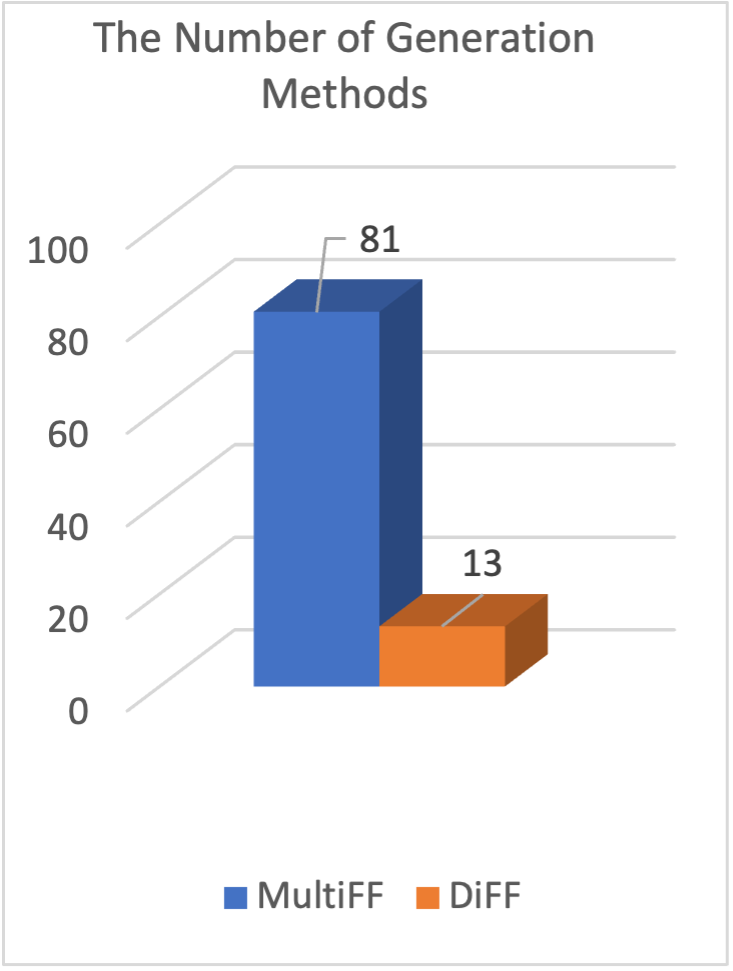}
\includegraphics[scale=0.45]{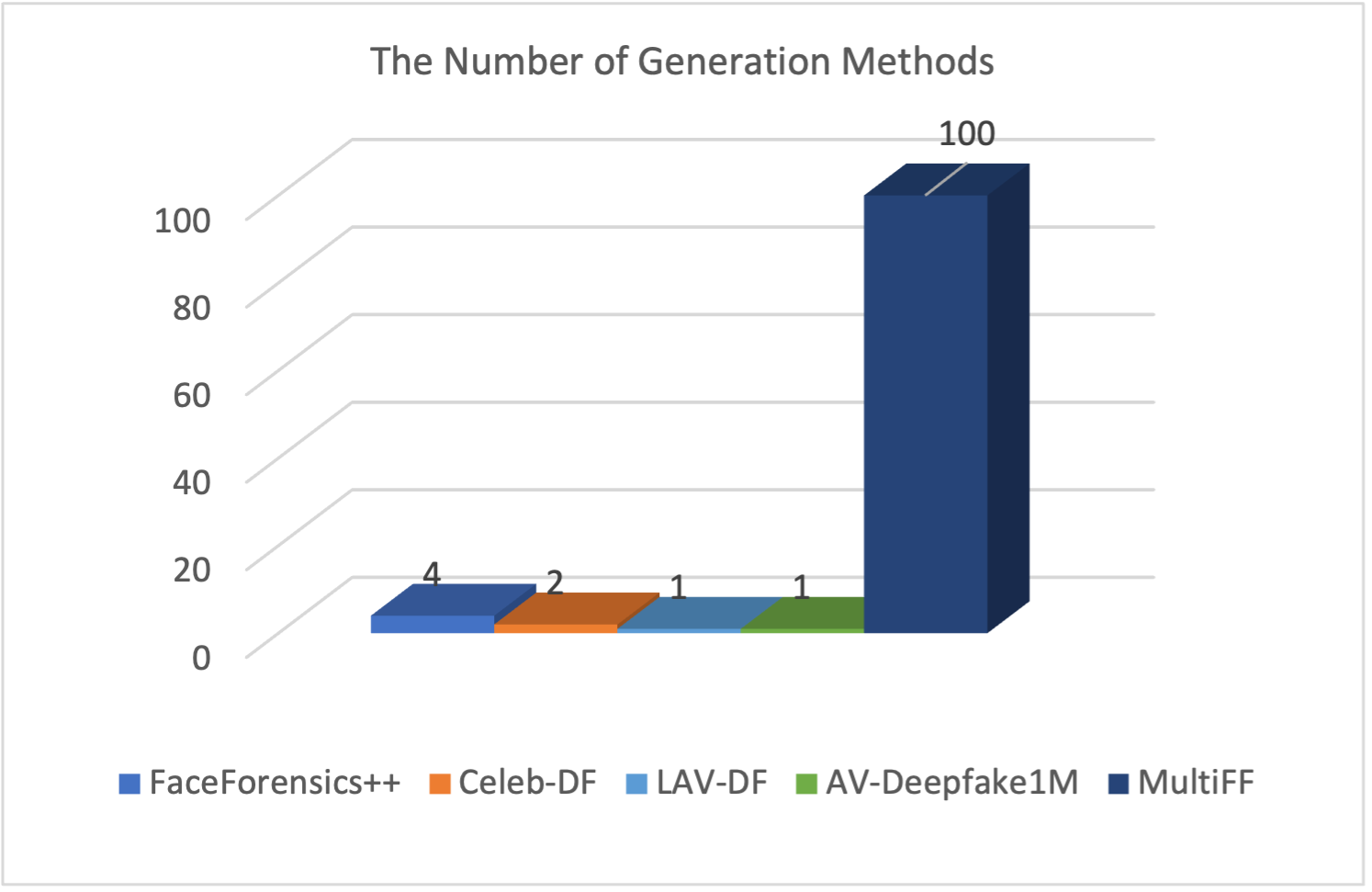}
\caption{The number of generation methods in MultiFFI (left) and MultiFFV (right), respectively.}
\label{fig:2_1}       
\end{figure}

\begin{table*}[ht]
    \centering
    \begin{tabular}{l|c|c|c|c}
    \toprule
        MultiFF & Real images & Forged images & Real audio-video & Forged audio-video \\
        \midrule
        training set & 99386 & 425043 & 68035 & 173955\\
        validation set & 59082 & 88281 & 27514 & 51994 \\
        public testing set & 77602 & 96785 & 44089 & 128382 \\
        hidden testing set & 156720 & 176129 & 33780 & 101135 \\
        \bottomrule
    \end{tabular}
    \caption{The breakdown of number of MultiFFI and MultiFFV in the MultiFF dataset.}
    \label{tab:1}
\end{table*}

Since the forgery of faces may cause more threats to AI systems, in our challenge we focus more on the face region rather than the background areas. As a result, all images in our dataset are aligned and cropped to 512 $\times$ 512, where the ratio of face regions is about 0.6 $\sim$ 0.7. All frames in the audio-video are aligned and cropped to 384 $\times$ 384, where the ratio of face regions is about 0.4 $\sim$ 0.8. Moreover, we are also concerned about the generalization performance of the algorithm. Thus the testing sets contain new and unseen forgery types compared to the training and validation sets, in order to measure the generalization capability of forgery detection models.

%% file: sec/3_finalcopy.tex
\section{Challenge Setup}

\label{sec:challenge_setup}
\subsection{Organizers}
Our challenge was hosted in conjunction with Inclusion 2024. The organizers are Ant Group, China Society of Image and Graphics, Advanced Technology Exploration Community, Ant Security Lab, University of Science and Technology of China, Centre For Frontier AI Research, Alibaba Cloud, Hunan University, Sun Yat-sen University, Fudan University, Shanghai Jiao Tong University, National University of Singapore, Nanyang Technological University, Datawhale, Sunthy Cloud, etc. The technical program was hosted on the Kaggle platform \footnotemark[3].
\footnotetext[3]{https://www.kaggle.com/competitions/multi-ffdi}

\subsection{Processes}
Our challenge has two tracks including image forgery detection (Track 1) and audio-video forgery detection (Track 2). The whole challenge was divided into three phases, i,e. Phase 1, Phase 2, Phase 3.

In Phase 1, only the training and validation sets are released. The forgery detection model can only be trained on the training set, with ImageNet pre-trained model weights in Track 1 and pre-trained model weights in Track 2. Data from external sources are not allowed in the training process. However, some image processing methods, such as face detection and alignment, and face enhancement from the training dataset, are allowed to be used in the challenge. The validation set is also released, which can be used by the participants to improve the model performance and select the best model. Phase 1 lasted for about two months to provide enough time to perform model training and validation.

In Phase 2, the public testing set was released. Participants can directly submit the predicted score of the testing set to the platform and get immediate feedback on the evaluation scores twice a day. Phase 2 lasted for eight days to avoid over-fitting of the testing set.

The top 20 teams from the leaderboard during Phase 2 can advance to the final Phase 3. In Phase 3, codes and models are submitted together with technical reports, which are used to produce prediction scores on our hidden testing set. The final ranking will be based on the weighted score of the public testing set, the hidden testing set and the technical report, and the weights are 0.2, 0.6, and 0.2, respectively.

\subsection{Evaluation Methods}
For the performance evaluation, we mainly use the Area under the Curve (AUC) in both two tracks. AUC is defined as the area under the Receiver Operating Characteristic (ROC) curve, and the value range is generally between 0.5 and 1. To be specific, in our setting, True Positive (TP), True Negative (TN), False Positive (FP) and False Negative (FN) are defined as follows:

\begin{enumerate}
    \item TP: The forged images are recognized as forged images
    \item TN: The real images are recognized as real images
    \item FP: The real images are recognized as forged images
    \item FN: The forged images are recognized as real images
\end{enumerate}

With that, the True Positive Rate (TPR) and False Positive Rate (FPR) are defined in Equation (\ref{eq:1}) and Equation (\ref{eq:2}), respectively.

\begin{equation}
    TPR=\frac{TP}{(TP+FN)}
    \label{eq:1}
\end{equation}

\begin{equation}
    FPR=\frac{FP}{(FP+TN)}
    \label{eq:2}
\end{equation}

The ROC curve is essentially the TPR v.s. FPR curve and AUC is the area under this curve. To further assess and analyze the models, TPR at lower FPR, such as 1e-2, 5e-3, and 1e-3 will also be used as auxiliary metrics. However, the rankings will still be based on AUC. 

%% file: sec/4_solution.tex
\section{Solutions and Results}
Our challenge has attracted 1500 teams with valid submission counts. Our final validation set leaderboard has teams and the final test set leaderboard has teams. The top 20 teams are invited to the final phase, Phase 3 in each track.

In the following subsections, we will present the solutions of the top 3 teams in each track of our challenge.

\subsection{Track 1: Image Forgery Detection}
\subsubsection{Solution of the 1st Place}

\begin{itemize}
    \item Solution title: Towards Generalizable Deepfake Detection via Clustered and Adversarial Forgery Learning
    \item Team Name: JTGroup
    \item Team members: chxy95, fengpengli, highwayw, kahimwong, kemoulee, namecantbenull, rebeccaee, umlizheng, yiyayoo
\end{itemize}

\begin{figure*}[htbp]
\centering
\includegraphics[width=0.9\linewidth]{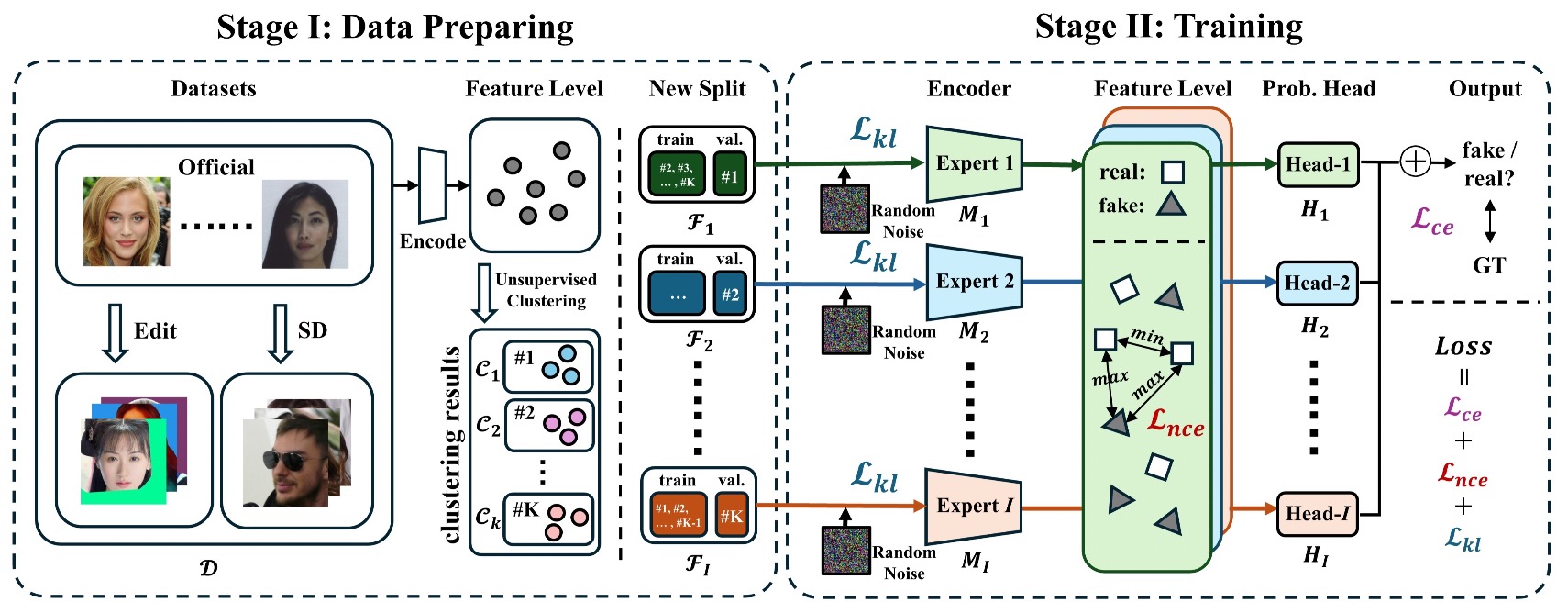}
\caption{Method overview of JTGroup.}
\label{fig:jtgroup_1}       
\end{figure*}

\paragraph{General Method Description}
The champion team proposed a generalized method for image forgery detection which can be categorized into two stages: (1) Data preparation, and (2) Training, with their fundamental processes illustrated in Fig. \ref{fig:jtgroup_1}.
\subparagraph{Data preparation}

\begin{figure*}[htbp]
\centering
\includegraphics[width=0.7
\textwidth]{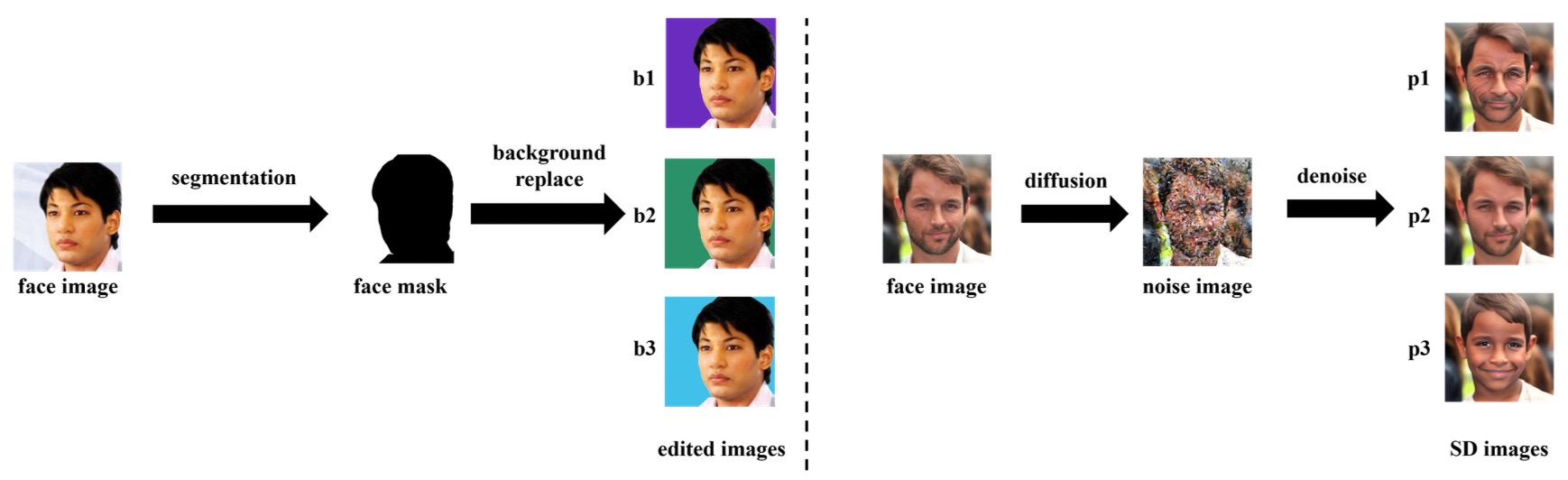}
\caption{Data preparation of JTGroup. The left side illustrates the image editing operations, while the right side demonstrates data generation using Stable Diffusion.}
\label{fig:jtgroup_2}       
\end{figure*}

To enhance model generalization and alleviate overfitting, the proposed solution expands the training dataset using a combination of image editing and Stable Diffusion (SD) \cite{ho2020denoising} techniques. As shown in Fig. \ref{fig:jtgroup_2}, the data generation process involves the following operations:
The first technique is image editing, which involves altering specific elements of the original images to create new variants. Initially, the team applied facial semantic segmentation to isolate the facial region and background. Once separated, the background is modified with different colors (e.g., purple, green, and blue), while preserving the original facial features. The second technique utilizes the SD model to generate new images from the original dataset. The resulting images reflect a wide range of styles and features, enriching the training set with diverse representations of both real and manipulated data.
\subparagraph{Data clustering}
The types of forgery encountered during testing often differ from those seen during training. This solution proposes a Data Clustering method to reallocate the training and validation datasets. The primary objective is to cluster the dataset according to forgery types, thereby simulating practical testing conditions where the model is exposed to a broader variety of unseen forgeries.

\begin{figure*}[htbp]
\centering
\includegraphics[width=0.7
\textwidth]{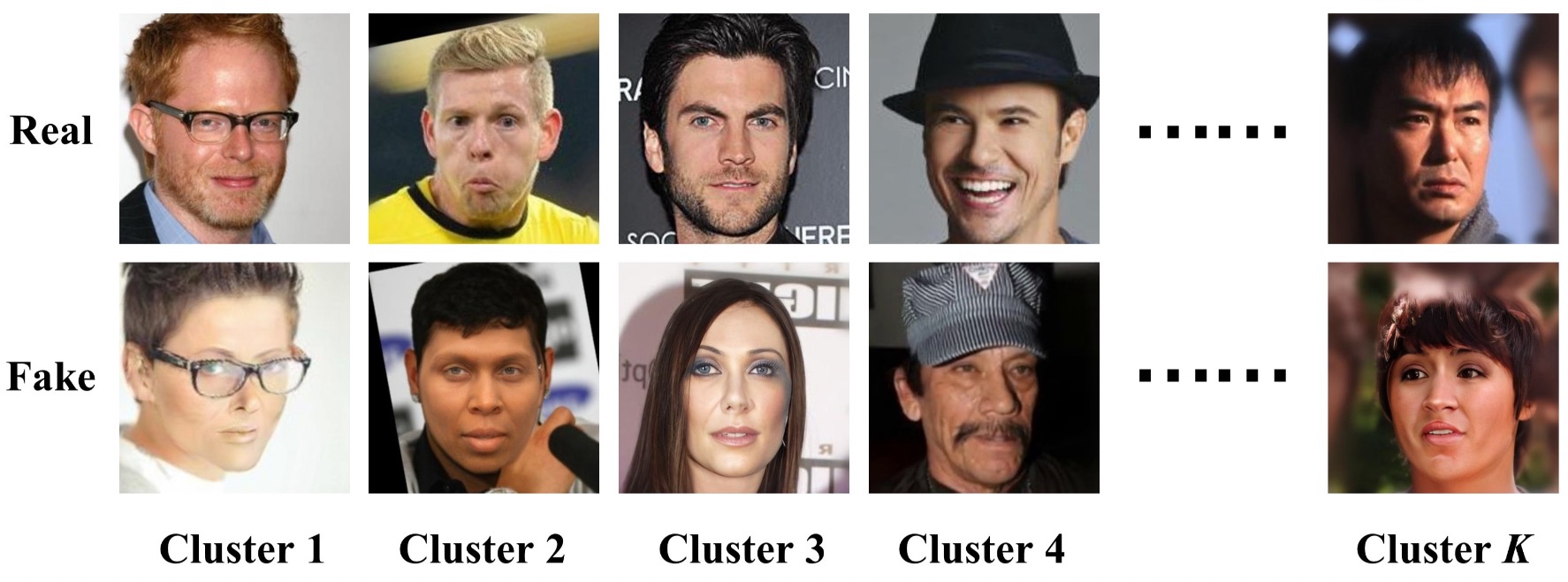}
\caption{Real and fake data after clustering. Clustering operates at the feature level, and its effectiveness may not be fully reflected at the pixel level.}
\label{fig:jtgroup_3}       
\end{figure*}

\subparagraph{Network Architecture}
The network architecture for the training stage is designed to ensure that the model can effectively generalize across different data distributions while remaining robust to forgery and adversarial attacks. As depicted in Fig. \ref{fig:jtgroup_1}, this architecture consists of several integral components, each with specific roles contributing to the overall model performance. The original images I from each fold are fed into trainable expert models $\{M_1(I;\theta_1), M_2(I;\theta_2), ..., M_I(I;\theta_I)\}$. Each expert model $M_i$, parameterized by $\theta_i$, is specialized to learn from the images within its respective fold. The output of each expert is a high-dimensional feature vector $v_i=M_i(I;\theta_i)\in \mathbb{R}^d$, which is used in two critical operations subsequently: first, it is employed in the calculation of the InfoNCE loss $\mathcal{L}_{NCE}$; second, it is passed through a probabilistic head to generate logits for the cross-entropy loss $\mathcal{L}_{CE}$.

\subparagraph{Training description}
During training, the team primarily utilized balanced sampling of positive and negative samples, the cosine annealing learning rate adjustment strategy, and exponential moving average (EMA) \cite{klinker2011exponential} weighted weight smoothing. The number of clusters K for the unsupervised clustering algorithm is set to 20 and there are $I=7$ expert models in the ensemble, each trained on a different fold created through the clustering process. The decay factor $\gamma$ for EMA is set to 0.995. The augmentations used include but are not limited to, JPEG and WebP compression, blur, Gaussian noise, random brightness, and grid distortion.
\subparagraph{Testing description}
During testing, the final prediction is determined by averaging the logits across all experts and applying the sigmoid function.

\paragraph{Generalization Analysis}
The champion team evaluated the performance of the proposed method on the MultiFFI dataset using cross-entropy loss and the AUC evaluation metric, as shown in Table \ref{tab:2}. It can be observed from the table that although the baselines (EfficientNet and ConvNeXt) perform well on the validation set, achieving AUC scores above 0.99, they do not generalize effectively to the public test set. The proposed framework introduces innovative clustering at the dataset level, making the corresponding validation sets more challenging. By incorporating clustering and adversarial optimization objectives, the learned forgery features exhibit enhanced generalization. To further evaluate proposed method in detecting different or unknown types of forgeries (such as face swapping, face reenactment, facial attribute editing, face synthesis, etc.), the team conduct additional experiments on the well-known datasets FaceForensics++ \cite{rossler2019faceforensics++}, DFDC \cite{dolhansky2020deepfake}, and DFD \cite{dufour2019google}. The experimental results, using AUC as the evaluation metric, are presented in Table \ref{tab:00}. State-of-the-art methods such as SBI \cite{shiohara2022detecting}, RECEE \cite{cao2022end}, and CFM \cite{luo2023beyond} are introduced for comparison with our method. To ensure a fair comparison, the team retrained all methods on the FaceForensics++ training set. Additionally, the champion solution from the 2019 DFDC competition (DFDC-1st-place) is included as a reference.

\begin{table*}[ht]
    \centering
    \resizebox{1.0\linewidth}{!}{
    \begin{tabular}{cccccccc}
    \toprule
        \multirow{2}{*}{Method} & \multirow{2}{*}{Venue} & \multicolumn{4}{c}{FaceForensics++} & \multirow{2}{*}{DFDC} & \multirow{2}{*}{DFD}\\ \cline{3-3} \cline{4-4} \cline{5-5} \cline{6-6}
         & & Deepfake & Face2Face & FaceSwap & NeuralTextures \\ \hline
    DFDC-1st-place&-&-&-&-&-&0.8130&0.7211\\ \hline
    SBI&CVPR'22&0.9993&0.9927&0.9953&0.9915&0.8251&0.8268\\
    RECEE&CVPR'22&0.9995&0.9920&0.9972&0.9959&0.6690&0.8687\\
    CFM&TIFS'23&0.9993&0.9923&0.9985&0.9924&0.8022&0.9123\\
    Baseline (EN-B4)&-&0.9990&0.9913&0.9962&0.9910&0.7863&0.8867\\
    Ours (EN-B4)&-&0.9998&0.9954&0.9992&0.9965&0.8292&0.9265\\
    \bottomrule
    \end{tabular}}
    \caption{Qualitative advantages of the proposed method for detecting defferent types of forgery attacks.}
    \label{tab:00}
\end{table*}

\begin{table*}[ht]
    \centering
    \resizebox{1.0\linewidth}{!}{
    \begin{tabular}{ccccccccccc}
    \toprule
    \multicolumn{2}{c}{Method} & \multicolumn{2}{c}{Official Train} & \multicolumn{2}{c}{Official Val} & \multicolumn{2}{c}{Split Train} & \multicolumn{2}{c}{Split Val} & Public Test \\
    \midrule
    Framework & Network & Loss & AUC & Loss & AUC & Loss & AUC & Loss & AUC & AUC \\
    \midrule
    \multirow{3}*{Baseline} & Efficient-B0 & 0.2583 & 0.9999 & 0.2637 & 0.9594 & - & - & - & - & 0.86379 \\
    ~ & Efficient-B4 & 0.2249 & 0.9999 & 0.2414 & 0.9939 & - & - & - & - & 0.92998 \\
    ~ & ConvNeXt-Tiny & 0.1539 & 0.9999 & 0.1626 & 0.9968 & - & - & - & - & - \\
    \multirow{5}*{JTGroup} & Efficient-B0 & - & - & - & - & 0.2671 & 0.9999 & 0.2781 & 0.9852 & 0.90767 \\
    ~ & Efficient-B4 & - & - & - & - & 0.2241 & 0.9999 & 0.2389 & 0.9977 & 0.94806 \\
    ~ & Efficient-B6 & - & - & - & - & 0.1910 & 0.9999 & 0.2059 & 0.9986 & 0.97547 \\
    ~ & ConvNeXt-Tiny & - & - & - & - & 0.1529 & 0.9999 & 0.1659 & 0.9985 & 0.97588 \\
    ~ & Final & \multicolumn{8}{c}{N/A} & 0.98051 \\
    \bottomrule
    \end{tabular}}
    \caption{Results of generalization evaluation of the proposed method on the official split dataset, our re-split dataset, and the public test set.}
    \label{tab:2}
\end{table*}

\subsubsection{Solution of the 2nd Place}

\begin{figure*}[htbp]
\centering
\includegraphics[width=0.7
\textwidth]{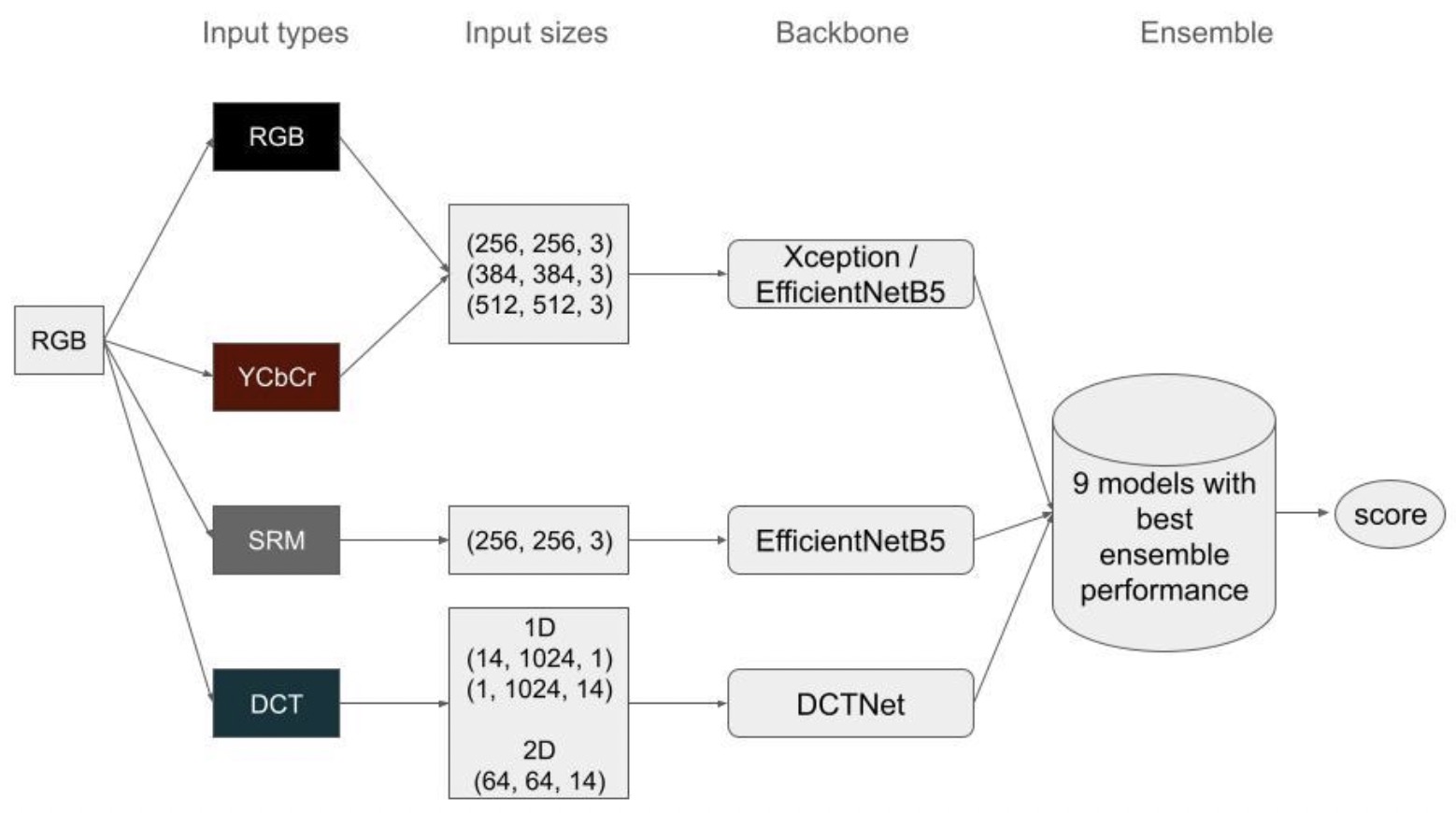}
\caption{Representative image of the Aegis method.}
\label{fig:aegis_1}       
\end{figure*}

\begin{itemize}
    \item Solution title: A Multi-Dimensional Method for Deepfake Detection
    \item Team Name: Aegis
    \item Team members: starethics
\end{itemize}
\paragraph{General Method Description}
To improve the generalization of forgery detection ability, the solution mainly focuses on four aspects: (1) data augmentation and synthesis, (2) model selection, (3) input modality selection, and (4) model fusion. The main process of the method is as follows, and this pipeline is shown in Figure \ref{fig:aegis_1}.

\subparagraph{Data augmentation and synthesis}
First, seven data augmentation methods are used to create more fake training data. The seven data augmentation methods are blur, gamma adjustment, hsv-based color adjustment, random crop, random noise, rotation, and horizontal flip. Second, with the hypothesis that synthesizing more deepfake data with more different methods might be able to increase the coverage of the training set and thus improve the model's performance, the team tried to synthesize more deepfake data with several methods (inswapper, SimSwap \cite{chen2020simswap}, E4S2024, Face-Adapter, Face X-ray, DiffFace, etc).

\subparagraph{Model selection}
Then, models are selected with different backbones, namely MobileNet, EfficientNet, Xception, Mobileone, Swin Transformer, etc, and the models are trained on the training set composed of the original data, the data augmentation, and synthesis results.

\subparagraph{Input Modality selection}
Different from most other traditional computer vision tasks which mainly focus on learning semantic features, deepfake detection needs to focus more on many non-semantic features/patterns. To help the model to better learn these features, the team designed different types of inputs, such as YCbCr (in this color space, a luma signal is isolated and can better represent the information of brightness distribution), SRM (a convolution filter which can be applied to the image and is able to help extract the noise pattern of images) and DCT (a format in 1D feature vector or 2D matrix).

\subparagraph{Model fusion}
After obtaining every single model, these models are evaluated on the official validation set and their self-made dataset. The team trained a small ensemble model with Attention and FC, which accepts prediction scores from every single model and outputs a final score. The ensemble model will perform better than simply averaging all scores.

\subsubsection{Solution of the 3rd Place}
\begin{figure*}[htbp]
\centering
\includegraphics[width=0.7
\textwidth]{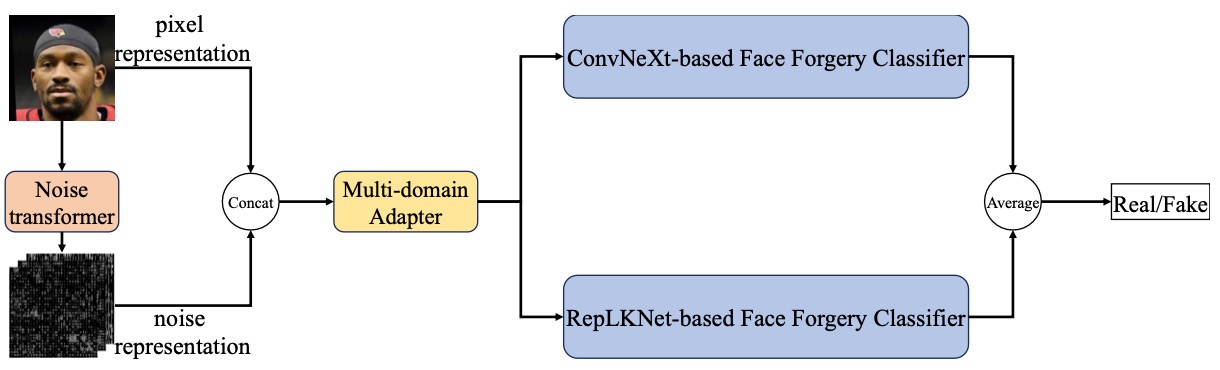}
\caption{The overall architecture of VisionRush multi-domain fusion and multi-model ensemble based face forgery detection method.}
\label{fig:visionrush_1}       
\end{figure*}

\begin{itemize}
    \item Solution title: Multi-domain Fusion and Multi-model Ensemble for Face Forgery Detection
    \item Team Name: VisionRush
    \item Team members: youwenwang01, zpp159541, qhukaggle, zhonghuazhao, tchj65539, Fieldhunter
\end{itemize}
\paragraph{General Method Description}
In this challenge, the 3rd team proposes a multi-domain fusion and multi-modal ensemble-based face forgery detection framework, as shown in Figure \ref{fig:visionrush_1}. The key design lies in two points: Firstly, they simultaneously utilize the pixel domain representation and noise domain representation of facial images as inputs. Secondly, they construct forgery classification models based on ConvNeXt and RepLKNet backbones respectively, and fuse the predicted forgery scores of the two models as the final result.

\begin{figure*}[ht]
\centering
\includegraphics[width=0.8
\textwidth]{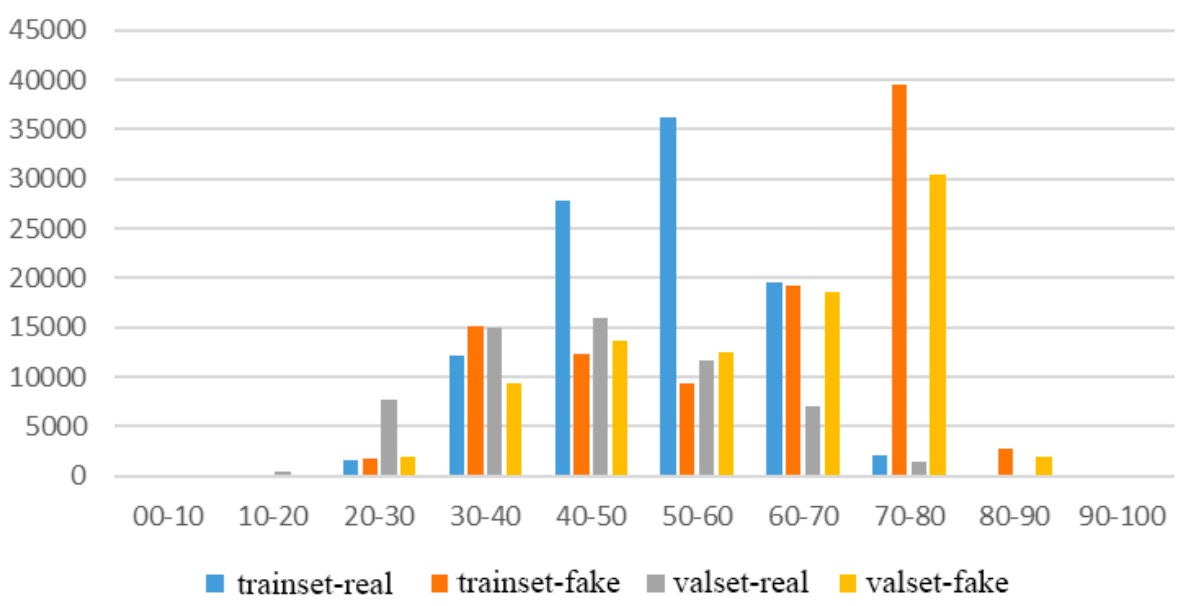}
\caption{Data distribution of real and fake images based on CenseoQoE.}
\label{fig:visionrush_2}       
\end{figure*}

\paragraph{Data augmentation}
\subparagraph{The visual quality degradation for fake data}
In the preliminary observation of the data, the team found that the distribution of visual quality of images is quite different. VisionRush initially performed a comprehensive evaluation of the competition dataset from the perspective of image quality. Employing the CenseoQoE-SDK, a sophisticated image and video quality assessment tool, they meticulously analyze the training and validation sets from the first phase. The CenseoQoE model assigns a predictive score ranging from 0\% to 100\% to each image, with higher scores indicating superior image quality. The analytical results for the training and validation sets are illustrated in Figure \ref{fig:visionrush_2}.

\begin{figure}[ht]
\centering
\includegraphics[scale=0.2]{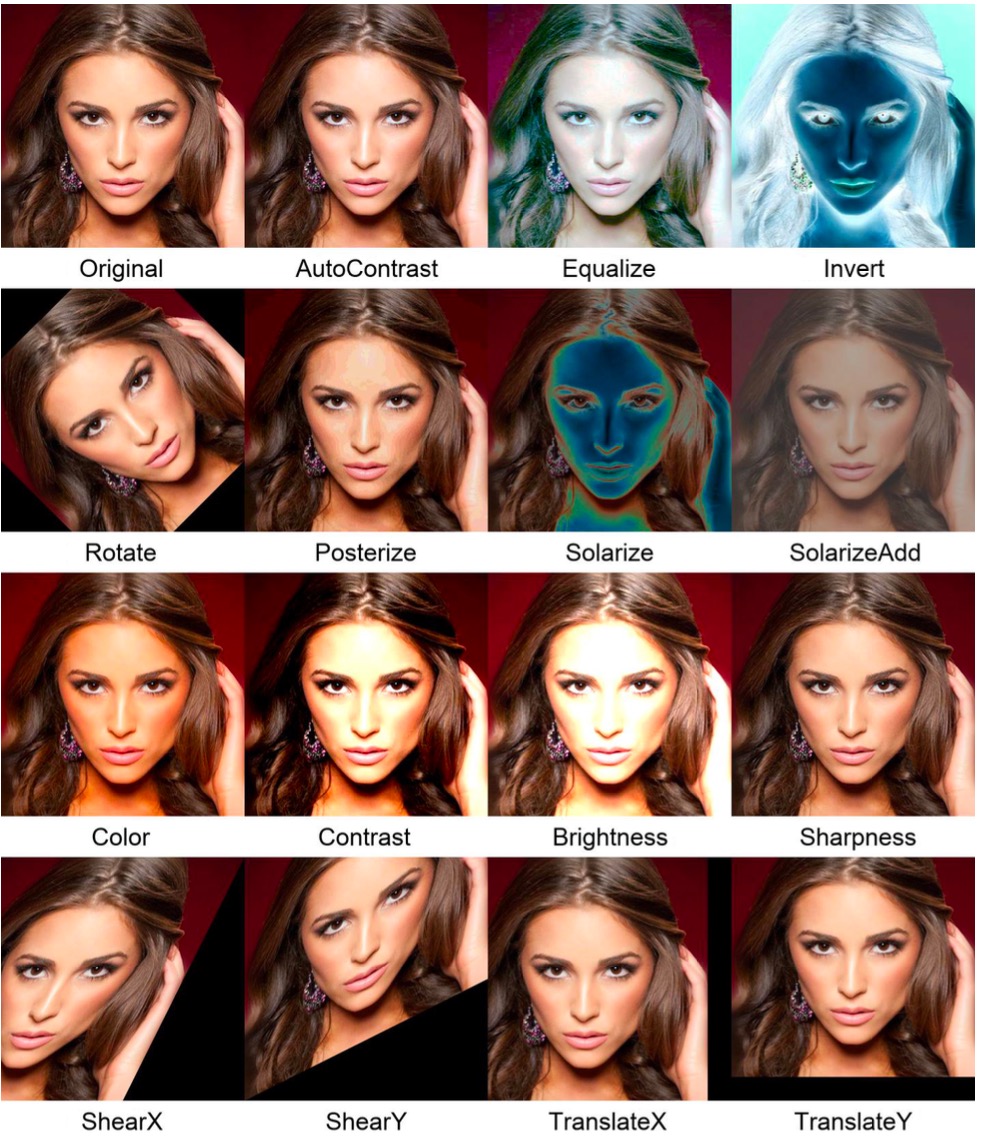}
\caption{Examples of 16 data augmentation effects.}
\label{fig:visionrush_3}       
\end{figure}

\subparagraph{The general augmentation for all data}
In addition to performing quality degradation operations on forged data, the team also applied general augmentation operations to all data during the preprocessing stage of training. Specifically, they follow the rand-m9-mstd0.5-inc1 configuration in RandAugment, which includes 15 different image processing operations such as contrast adjustment, histogram equalization, rotation transformation, and shear transformation, as illustrated in Figure \ref{fig:visionrush_3}. During model training, the system randomly selects and applies two strategies from these 15 options.

\paragraph{Implementation details}
The team trained ConvNeXt-based and RepLKNet-based real/fake binary classifiers respectively, with hyper-parameter settings detailed in Table \ref{tab:3}. For pre-training, they utilize publicly available weights trained on the ImageNet-1K dataset for each backbone. During training, the AdamW optimizer with a cosine annealing strategy is employed, where the learning rate gradually decreases from various initial values to 1e-6. The number of training epochs is set to 20. In addition, the team applied the Exponential Moving Average (EMA) technique to obtain more robust and generalized model weights in the training stage. During testing, they first set the image resolution to 512x512 and leverage multiple additional data perspectives to further improve inference performance, including $90^{\circ}$, $180^{\circ}$, and $270^{\circ}$ rotations, as well as horizontal and vertical flips. Then we average the predicted probabilities of the two classifiers as the final result.

\begin{table*}[ht]
    \centering
    \begin{tabular}{ccccc}
    \toprule
         Backbone & Batch Size & Input Resolution & Initial Learning Rate\\
         \midrule
         ConvNeXt & 192 & 384x384 & 1e-4 \\
         \midrule
         RepLKNet & 128 & 384x384 & 1e-4 \\
         \bottomrule
    \end{tabular}
    \caption{Training settings for different models.}
    \label{tab:3}
\end{table*}

\paragraph{Generalization Analysis}
To verify the generalization ability of the model, the team collected a large amount of data for testing, including synthetic data collected from various AIGC platforms and deep synthesis tools on the Internet, and some real data selected from academic datasets. The test results are shown in Table \ref{tab:4}. It can be seen that the method achieves high performance on data generated by all platforms and tools, demonstrating strong generalization ability.

\begin{table}[ht]
    \centering
    \resizebox{0.45\textwidth}{!}{
    \begin{tabular}{cccc}
    \toprule
         Platform/Tool/Dataset & Total Number & Correct Number & Recall \\
         \midrule
         \multicolumn{4}{c}{Fake data}\\ 
         \midrule
         Draft & 384 & 384 & 100\%\\
         \midrule
         NetEase\_AI\_Design\_Workshop & 932 & 907 & 97.31\%\\
         \midrule
         JourneyArtAI & 2054 & 2042 & 99.41\%\\
         \midrule
         liblibai & 983 & 983 & 100\%\\
         \midrule
         miaohua & 369 & 368 & 99.72\%\\
         \midrule
         6pen.art & 289 & 288 & 99.65\%\\
         \midrule
         artguru & 356 & 354 & 99.43\%\\
         \midrule
         imagine\_ai & 386 & 384 & 99.48\%\\
         \midrule
         promptthunt & 400 & 400 & 100\%\\
         \midrule
         WomboVERSE & 406 & 377 & 92.85\%\\
         \midrule
         shedevrum & 429 & 418 & 97.43\%\\
         \midrule
         wujieAI & 492 & 492 & 100\%\\
         \midrule
         diffusionbee & 384 & 377 & 98.17\%\\
         \midrule
         eSheep & 460 & 454 & 98.69\%\\
         \midrule
         MewXAI & 460 & 452 & 98.26\%\\
         \midrule
         XingZhiHuiHua & 6493 & 6438 & 99.15\%\\
         \midrule
         XiaoKuAI & 447 & 447 & 100\%\\
         \midrule
         chushouAI & 428 & 419 & 97.89\%\\
         \midrule
         Roop & 132 & 132 & 100\%\\
         \midrule
         FaceFusion & 203 & 202 & 99.51\%\\
         \midrule
         DoFaker & 193 & 190 & 98.45\%\\
         \midrule
         Total & 16680 & 16508 & 98.97\%\\
         \midrule
         \multicolumn{4}{c}{Real data}\\
         \midrule
         Glint360K & 5000 & 4676 & 93.52\%\\
         \midrule
         FFHQ & 5000 & 4832 & 96.64\%\\
         \midrule
         COCO2017 & 5000 & 4725 & 94.50\%\\
         \midrule
         Total & 15000 & 14233 & 94.89\%\\
         \bottomrule
    \end{tabular}}
    \caption{Generalization evaluation of 3rd method.}
    \label{tab:4}
\end{table}

\subsection{Track 2: Audio-Video Forgery Detection}

\subsubsection{Solution of the 1st Place}
\begin{itemize}
    \item Solution title: Audio-visual Deepfake Detection via spectrum and spatial joint learning

    \item Team Name: chuxiliyixiaosa
    \item Team members: chuxiliyixiaosa
\end{itemize}

\begin{figure*}[htbp]
\centering
\includegraphics[width=0.6
\textwidth]{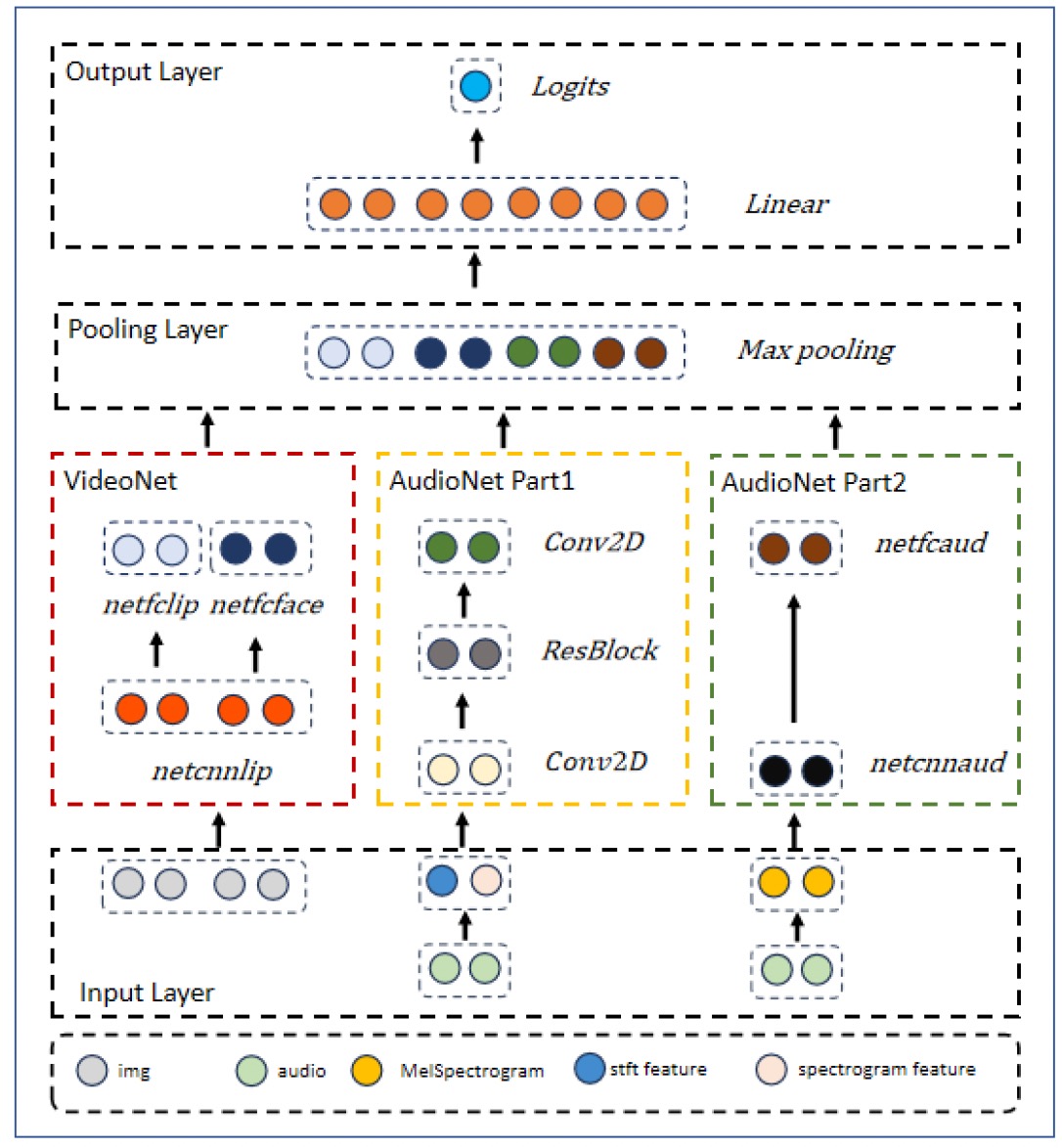}
\caption{Method Overview}
\label{fig:1}       
\end{figure*}

\paragraph{General Method Description}
To boost the model's forgery detection capabilities, the proposed solution leverages joint video-audio learning using SyncNet as the backbone structure. As illustrated in Fig.\ref{fig:1}, the approach involves simultaneous processing of video and audio inputs through joint learning, exploiting temporal and spectral features from both modalities. The video component utilizes VideoNet to extract sequential frame features, while the audio component employs Short-Time Fourier Transform (STFT) and Mel-spectrograms to capture audio features. Notably, the architecture includes dedicated modules for lip and face feature extraction, enabling the model to focus on subtle inconsistencies between audio and video inputs. The feature outputs are then combined through a fully connected layer, pooling features from both modalities to generate a probability score indicating the input's authenticity. This approach emphasizes capturing minute differences in audio-visual data, enhancing the model's ability to detect various deepfake operations.

\subparagraph{Data augmentation and synthesis}
Data augmentation plays a crucial role in enhancing the diversity and robustness of the training dataset. To achieve this, the model employs a mixup strategy during data loading, randomly concatenating two real video segments with a 40\% probability. This approach not only increases the diversity of real samples but also helps balance the distribution of real and fake videos in the dataset. Furthermore, when processing videos, the model limits the maximum number of frames to 900 consecutive frames, ensuring that sufficient information is preserved to learn rich temporal features while preventing memory overflow. Audio data is sampled at a 16kHz rate, aligned with video frames, and synchronized between the two modalities. This strategy effectively extracts useful information from the data, boosting the model's performance. Other data training parameters include: Image data is resized to a uniform size of (284, 284) and then normalized by dividing by 255.

\subparagraph{Network Architecture}
The model architecture plays a crucial role in achieving high performance in deepfake detection. This architecture is based on SyncNet, which enables joint learning of video and audio modalities. The model constructs an efficient detection framework by extracting audio and video features, combining Short-Time Fourier Transform (STFT) and Mel-spectrograms. The video processing module, netcnnlip, extracts deep features from adjacent frames using 3D convolution, and the output is fed into netfclip and netfcface modules, which focus on lip and face feature extraction, respectively. Audio processing is divided into two parts: Audio Part 1 computes STFT and spectrograms, while Audio Part 2 computes Mel-spectrograms, extracting audio features. The alignment of audio and video inputs ensures that the model can effectively learn the relationship between the two modalities, which is crucial for identifying deepfakes that may exhibit subtle differences in lip movement and speech. Max Pooling is applied to the last dimension of the tensors video out1, video out2, audio out1, and audio out2. The pooled tensors are concatenated and passed through a fully connected layer to obtain a probability value. This value is then compared with the labels using BCEWithLogitsLoss. The architecture employs a series of pooling operations to integrate features from different modules, including netfclip, netfcface, and netfcaud. The outputs of all modules are connected after max pooling, and the final output is passed through a fully connected layer to produce a probability value.

\subparagraph{Training description}
The training process of the deepfake detection model is optimized for efficiency and effectiveness. By leveraging the Distributed Data Parallel (DDP) method, the model is trained on multiple GPUs, accelerating the training time. Automatic Mixed Precision (AMP) is used to conserve GPU memory usage while accelerating computations.

\subparagraph{Testing description}
During the testing phase, the model is evaluated on a separate test set to assess its generalization ability.
\subsubsection{Solution of the 2nd Place}
\begin{itemize}
    \item Solution title: The Solution of Team ShuKing for Deepfake Video Detection.
    \item Team Name: ShuKing
    \item Team members:  Jack Hong (jaaackhong@gmail.com)
\end{itemize}

\begin{figure*}[htbp]
\centering
\includegraphics[width=0.6
\textwidth]{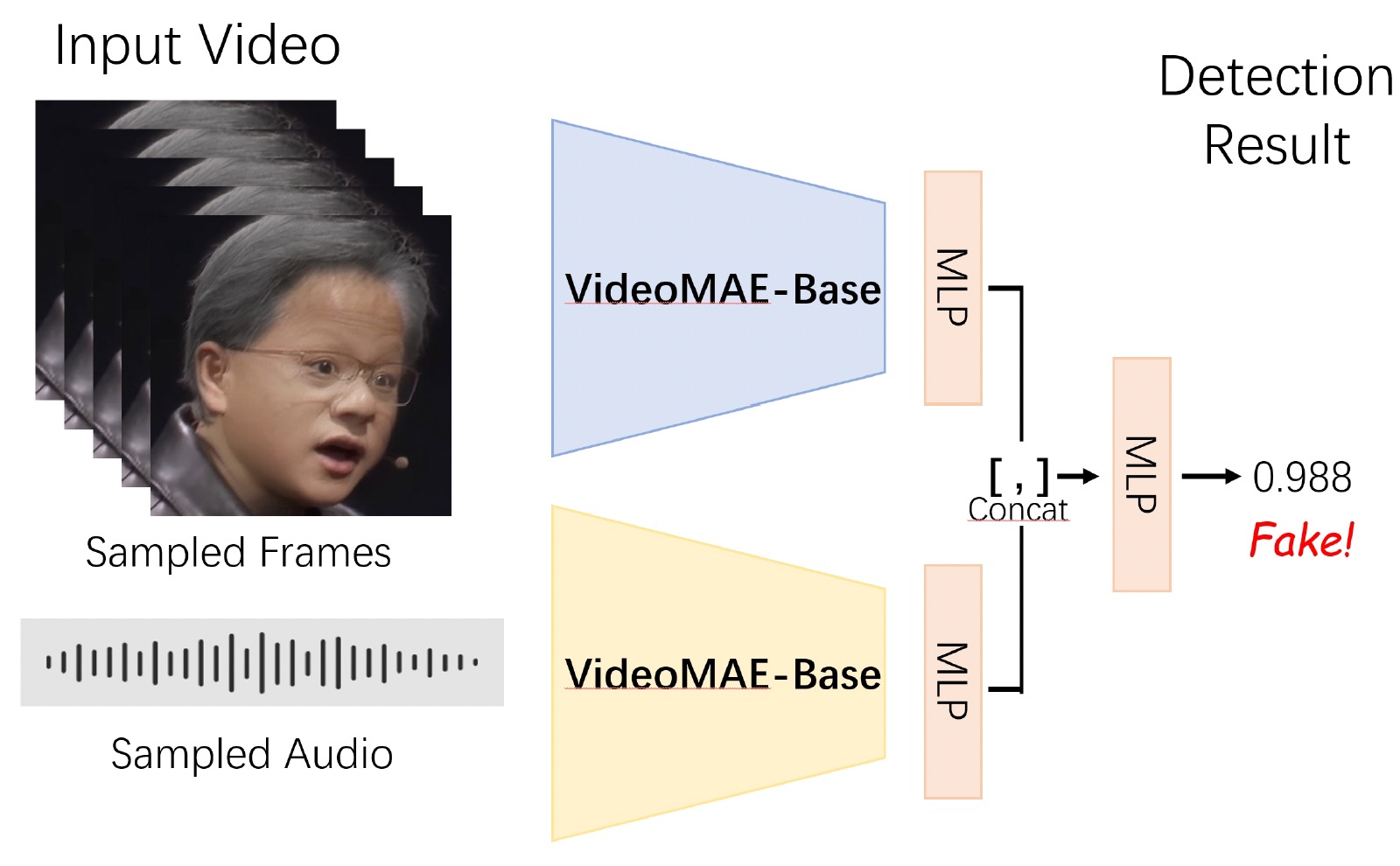}
\caption{Structure of the main model.}
\label{fig:2}       
\end{figure*}

\paragraph{General Method Description}
The deepfake video detection solution proposed by team ShuKing adopts a comprehensive approach to extracting both video and audio features. Fig.\ref{fig:2} illustrates the overall architecture of this approach. First, the team utilizes an advanced video foundation model, VideoMAE-Base, to extract high-level semantic features from the video, generating feature maps that capture object co-occurrence and contextual relationships within the video. By employing mean spatial and temporal pooling technology, the model can aggregate spatial and temporal information from the entire video, thereby identifying the overall patterns and structures of deepfake videos. On the audio side, the team converts the audio signal into Mel-spectrograms and uses another VideoMAE-Base model to extract audio features. This bimodal feature extraction approach ensures that the model considers both visual and auditory elements when distinguishing between real and AI-generated content. Finally, the team employs several MLP layers as the discriminator. This discriminator processes the extracted video and audio features to make the final prediction, ensuring a comprehensive analysis of both visual and auditory elements. The innovative aspect of this approach lies in its simultaneous analysis of video and audio features, which enhances the model's detection capabilities.

\subparagraph{Data augmentation and synthesis}
In terms of data augmentation, the ShuKing team employed a range of techniques to enhance the model's robustness and generalization ability. During training, the team implemented standard data augmentation strategies such as random scaling, cropping, and flipping. Additionally, the team introduced more advanced augmentation strategies, such as randomly sampling images and audio from different time points in the video to create diverse training samples. This temporal variation allowed the model to learn features from different video segments. Furthermore, the team occasionally replaced the original audio with audio from different videos, further increasing the diversity of the training data. This approach not only improved the model's adaptability to changes in audio content but also enhanced its ability to distinguish between real and AI-generated content. Through these data augmentation techniques, the model demonstrated stronger adaptability and accuracy when faced with different types of deepfake videos.

\subparagraph{Training description}
During the training stage, the ShuKing team used VideoMAE-Base as the base model and performed pre-training on the Kinetics-400K dataset to leverage the advantages of transfer learning. Each video was randomly sampled at 16 frames, with a frame rate of 4 FPS, to ensure that the model could capture the dynamic information in the videos. The learning rate was set to 5e-5, and the training process used 8 NVIDIA A100 GPUs, with each GPU processing a batch of 12 videos. The input videos were resized to a resolution of 224×224 pixels, and the entire training process lasted for 20 epochs. Through these strategies, the model was thoroughly trained on diverse data to improve its performance in real-world scenarios.

\subparagraph{Testing description}
During testing, the input videos are resized to 224x224 pixels. Each video is uniformly sampled at 16 frames, with a frame rate of 4 FPS. For videos exceeding 4 seconds, the team splits them into multiple 4-second segments, evaluates each segment separately, and selects the highest score as the final result. Additionally, the team employs the model soup technique, which averages the parameters from multiple trained models to enhance generalization and overall performance.

\subsubsection{Solution of the 3rd Place}
\begin{itemize}
    \item Solution title: Deepfake Audio-Video Detection Via MFCC Features
    \item Team Name: The Illusion Hunters
    \item Team members: JinXiaoxu, ZiyuXue, mppsk0

\end{itemize}
\begin{figure*}[htbp]
\centering
\includegraphics[width=0.9
\textwidth]{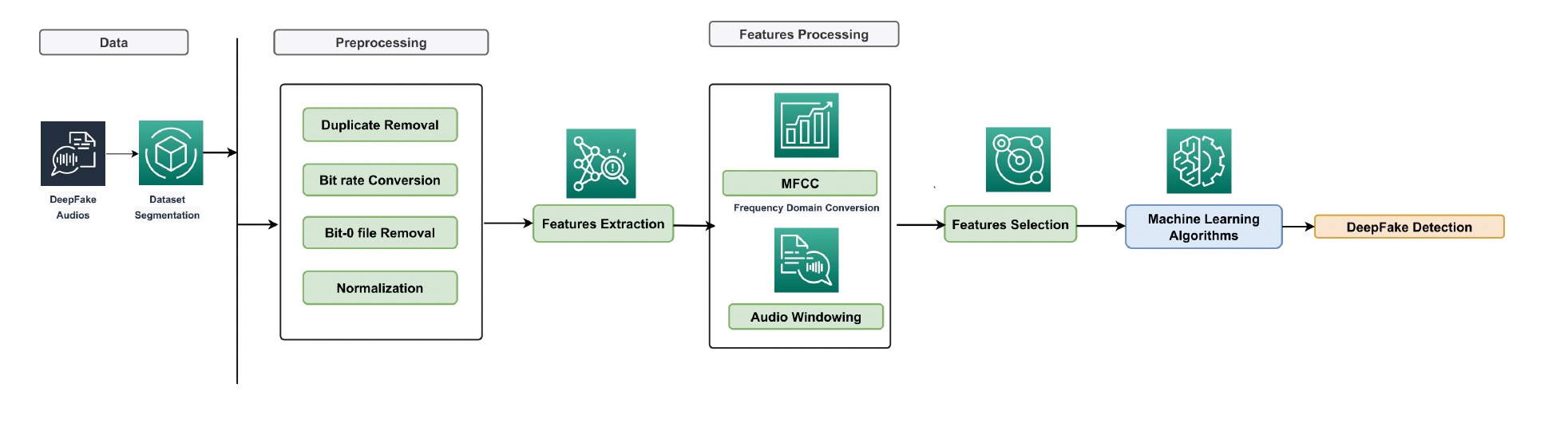}
\caption{Graphical representation for detection of deepfake audios.}
\label{fig:3}       
\end{figure*}

\paragraph{General Method Description}
The team "The Illusion Hunters" employs the Mel-Frequency Cepstral Coefficients (MFCCs) technique to extract useful features from the audio in video. Fig.\ref{fig:3} shows the complete process from data preprocessing, and feature extraction to feature processing and detection of deepfake audio.
Specifically, the MFCC features are computed using the MFCC function from the librosa library, and the arithmetic mean of these features is returned. Subsequently, a Support Vector Machine (SVM) is used for classification to determine whether the audio-visual content is authentic or manipulated. The core of this approach lies in the effective detection of deepfake content through the extraction and classification of audio features. Notably, this method avoids the use of complex deep learning models or pre-trained networks, opting instead for a traditional machine learning approach based on MFCC features. This choice results in a lower model complexity, enabling rapid training and deployment.

\subparagraph{Data augmentation and synthesis}
In terms of data preprocessing, the team extracts audio from video files and saves it in WAV format, laying the foundation for subsequent feature extraction.

\subparagraph{Training description}
During the training phase, the team sets the number of extracted Mel-Frequency Cepstral Coefficients (MFCCs) to 13 by default, the FFT window size to 2048, and the window hop size to 512. The team uses the librosa library to load the audio files compute their MFCC features, and return the mean of these features. Additionally, the team applies the StandardScaler to normalize the features, ensuring that they have a mean of 0 and a variance of 1, which accelerates convergence. For classification, the team employs a linear kernel Support Vector Machine (SVM).

\subparagraph{Testing description}
During the testing phase, the team's model uses standardized MFCC features as input, ensuring the reliability of the test results.

%% file: sec/5_discussions.tex
\section{Discussions}



\begin{figure}[htbp]
\centering
\includegraphics[scale=0.6]{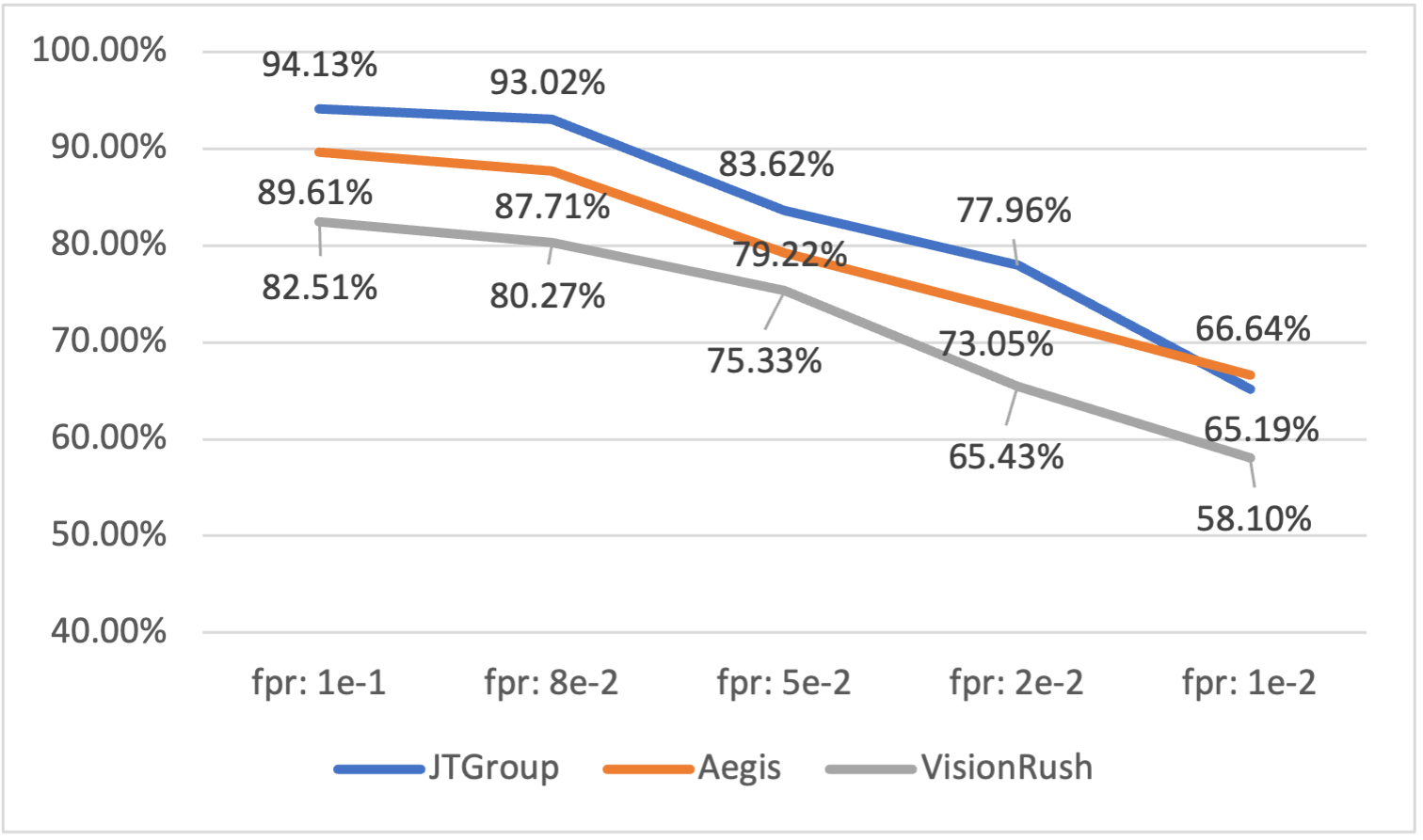}
\caption{TPR and FPR of top 3 teams in terms of unseen types of forgery on the test set in MultiFFI.}
\label{fig:5_1}       
\end{figure}

The technical reports indicate a prevalent use of data augmentation and data extension techniques in most submitted solutions. Participants are also exploring modeling approaches for unseen forgery types, feature and data representation, and model ensembling strategies. Although these efforts have led to relatively high area under the curve (AUC) scores, the challenges associated with image and audio-video forgery detection remain unresolved. To underscore this point, we conducted an in-depth analysis of the submitted solutions, with the findings presented Tables \ref{fig:5_1} and \ref{tab:5_2}.

A considerable performance disparity exists which is particularly pronounced when the false positive rate (FPR) is low. This performance gap can reach as much as 50\%. Furthermore, the true positive rate (TPR) at low FPR levels is suboptimal. The FPR metric is critical as it quantifies the rate at which authentic images are misclassified as forgeries, leading to an unnecessary inconvenience for users. In practical applications, where the user base is typically large, maintaining a minimal disturbance rate, represented by an FPR of 1/1000 or even 1/10000, is imperative. However, at such stringent FPR thresholds, the TPR, which indicates the effectiveness of correctly identifying forged images, is not yet sufficient for usability.

The analysis highlights the necessity for continued research efforts to narrow the performance gap between familiar and novel forgery types. Additionally, advancements are needed to enhance TPR at low FPRs, a challenge that warrants significant research focus.


\begin{table}[h!]
    \centering
    \resizebox{0.45\textwidth}{!}{
    \begin{tabular}{|l|ccccc|}
    \hline
    \diagbox{Team}{tpr}{fpr} & $1e-1$ &$8e-2$ &$5e-2$ & $2e-2$ & $1e-2$\\
    \hline
    chuxiliyixiaosa & 53.38\% & 47.98\% & 37.49\% & 22.93\%& 15.53\%\\
    ShuKing & 41.30\% & 38.76\% & 34.32\% & 27.17\% & 22.80\%\\
    The Illusion Hunters & 15.99\% & 12.09\% & 6.89\% & 2.94\%& 2.03\%\\
    \hline
    \end{tabular}}
    \caption{TPR and FPR of Top 3 teams in terms of unseen types of forgery on the test set in MultiFFV.}
    \label{tab:5_2}
\end{table}

%% file: sec/6_conclusion.tex
\section{Conclusion and Future Work}

In the realm of Global Multimedia Deepfake Detection, the challenge of multi-forgery detection has been formulated with precision and depth. This competition has illuminated several innovative approaches to tackling this complex problem, exemplified by the unforeseen efficacy of data-centric strategies, the development of simulations for previously unencountered types of forgeries, and the deployment of diverse model architectures offering unique inductive biases. The initiatives undertaken by the top three performing teams in each track of the competition reflect a spectrum of technical methodologies. These teams have employed cutting-edge data manipulation techniques and model training strategies to enhance the accuracy and reliability of deepfake detection systems. Their collective efforts underscore the importance of robust datasets and versatile models in navigating the intricate landscape of digital forgery detection.

To further advance research in this field, we have made available the MultiFF dataset to the wider research community. This dataset serves as a valuable resource for the testing and development of novel detection methodologies, offering a simulated environment that closely mirrors the adversarial and dynamic nature of real-world deepfake challenges.

The competition stands as a testament to the ongoing efforts to evaluate and improve the creation and detection of deepfakes in complex environments. It highlights the necessity of continual adaptation and innovation in methodologies to keep pace with the evolving threats posed by digital forgeries. Through collaborative efforts, the research community can better understand the intricacies of deepfake detection and develop strategies that ensure the integrity and authenticity of multimedia content in an increasingly digital world.

While this competition strives to simulate real-world deepfake attack scenarios as closely as possible, it neglects the exploration of interpretability in detection results. 
Existing academic research has investigated the interpretability of face deepfake detection, including single-face deepfake localization \cite{miao2023multi}, multi-face deepfake localization \cite{miao2024mixture}, and audio-visual deepfake temporal localization \cite{zhang2024mfms}. These studies offer a wealth of evidence for deepfake detection beyond the simplistic real/fake classification task. Consequently, we plan to incorporate forgery localization labels in the future to advance the development of interpretability in deepfake detection tasks. 